\documentclass[11pt,a4paper]{article}
\usepackage[margin=1.6cm,top=2cm,bottom=2cm]{geometry}
\usepackage[T1]{fontenc}
\usepackage{lmodern,graphicx,booktabs,amsmath,amssymb,tabularx,array,multirow,needspace}
\usepackage[font=small,labelfont=bf]{caption}
\usepackage[super,sort&compress]{natbib}
\usepackage[hidelinks]{hyperref}
\newcommand{\doi}[1]{\href{https://doi.org/#1}{doi:\nolinkurl{#1}}}

\title{Streaming artificial intelligence for cross-centre analysis of robotic prostatectomy workflows}
\author{\begin{tabular}{c}
Yang Liu$^{1,\dagger}$, Xianglong Shi$^{5,\dagger}$, Alexandre Calado$^{2}$, Yang Li$^{1}$,\\
Samy Cheik Youssef$^{3}$, Qiancheng Zhao$^{4}$, Armando Stabile$^{2}$, Sirui Zhao$^{5}$,\\
Enhong Chen$^{5}$, Cheng Shen$^{4}$, Xuesong Li$^{4}$, Hui Jiang$^{4,*}$,\\
Giorgio Gandaglia$^{2}$, Alberto Briganti$^{2}$, Francesco Montorsi$^{2,*}$, Benjamin Challacombe$^{3}$,\\
Jonathan Noel$^{3}$, Christian Brown$^{3}$, Paul Cathcart$^{3}$, Prokar Dasgupta$^{1,3}$,\\
Alejandro Granados$^{1}$, Nicholas Raison$^{1,3,*}$, S\'ebastien Ourselin$^{1}$
\end{tabular}}
\date{}
\begin{document}
\maketitle
\begin{small}
$^{1}$School of Biomedical Engineering \& Imaging Sciences, King's College London, London, UK.\\
$^{2}$Unit of Urology, Division of Oncology, IRCCS Ospedale San Raffaele, Milan, Italy.\\
$^{3}$Department of Urology, Guy's and St Thomas' NHS Foundation Trust, London, UK.\\
$^{4}$Department of Urology, Peking University First Hospital, Beijing, China.\\
$^{5}$University of Science and Technology of China, Hefei, China.\\
$^{\dagger}$Equal contribution.\\
$^{*}$Corresponding authors.
\end{small}

\begin{abstract}
Surgical phase recognition can turn robot-assisted radical prostatectomy videos into timelines for training, review and intraoperative assistance. They could also show how workflows differ between centres, which remains unclear. We previously developed SKiT, a streaming model that outputs the current phase while a fixed-size memory retains key information at constant per-frame cost. Here we trained SKiT separately in London, Milan and Beijing and compared the centres through annotations, cross-centre testing and model memory. Annotated phase order differed between centres. The Milan and Beijing models reached 87.14\% and 89.35\% frame accuracy at home but 72.27\% and 62.66\% on nine London recordings. Beijing was stronger in bladder mobilisation and Milan in posterior bladder-neck dissection. Averaging their probabilities recovered Beijing's strength but not Milan's. Compact eight-channel models retained frames from particular phases, in an order consistent only in London. Phase-level evaluation and traceable memory provide a basis for cross-centre analysis of surgical workflow.
\end{abstract}

\section*{Introduction}
Surgical video provides a detailed record of operative procedures, yet analysis remains largely dependent on time-consuming manual review. Accurately identifying the different phases of an operation, their sequence and how the procedure progresses through them is essential to leveraging the wealth of information contained within surgical recordings. Automated video analysis offers the potential to transform recordings into structured, clinically meaningful data to support surgical training, performance assessment, postoperative review and context-aware intraoperative assistance \cite{maierhein2017,maierhein2022}. For robot-assisted radical prostatectomy (RARP), the goal is to turn long recordings into navigable phase timelines that retain information about operative context. RARP phase recognition has been studied across surgeons, robotic platforms and institutions\cite{konnai2026,ueki2025,zhao2025rarp}, and recognition-derived timing has been used for skill evaluation\cite{zhao2025rarp}. Importantly, recognising surgical phases in real time, rather than retrospectively, is essential for enabling intraoperative applications. Although RARP follows a relatively standardised sequence of operative steps, differences in surgical techniques, equipment and operating environments between institutions present challenges for AI models. Developing models capable of accurate, real-time phase recognition that generalise across different centres is therefore essential for the wider clinical application of surgical AI. In gastric bypass, workflow models proved difficult to transfer between centres\cite{lavanchy2024}. Particularly when models are applied across different centres, overall phase recognition accuracy may conceal important errors in the identification of individual surgical phases. A model may achieve high overall accuracy while incorrectly identifying specific phases or their boundaries, resulting in inaccurate estimates of their duration \cite{reinke2024}. Evaluating the generalisability of surgical AI models therefore requires not only an assessment of overall recognition accuracy, but also an understanding of their performance across individual phases and the reliability of the surgical workflow measurements they produce.

Real-time surgical phase recognition presents an additional challenge, as predictions must be made online, using only information available up to that point in the operation. While recent video frames provide information about the current surgical activity, understanding the broader operative context may require information from earlier stages of the procedure. Temporal memory networks \cite{jin2021} and long-video transformers \cite{liu2025lovit} have been developed to incorporate historical information into surgical workflow recognition. We previously developed SKiT \cite{liu2023skit}, a streaming surgical phase recognition model that retains relevant historical information in a fixed-size memory, enabling real-time predictions without increasing computational requirements as the operation progresses. Although SKiT demonstrated accurate and computationally efficient phase recognition, its ability to generalise across different surgical centres has not been evaluated. Furthermore, while the model retains information from earlier stages of the operation, the extent to which this memory captures clinically meaningful surgical contexts remains unclear.

\begin{figure}[htbp]
\centering\includegraphics[width=\linewidth]{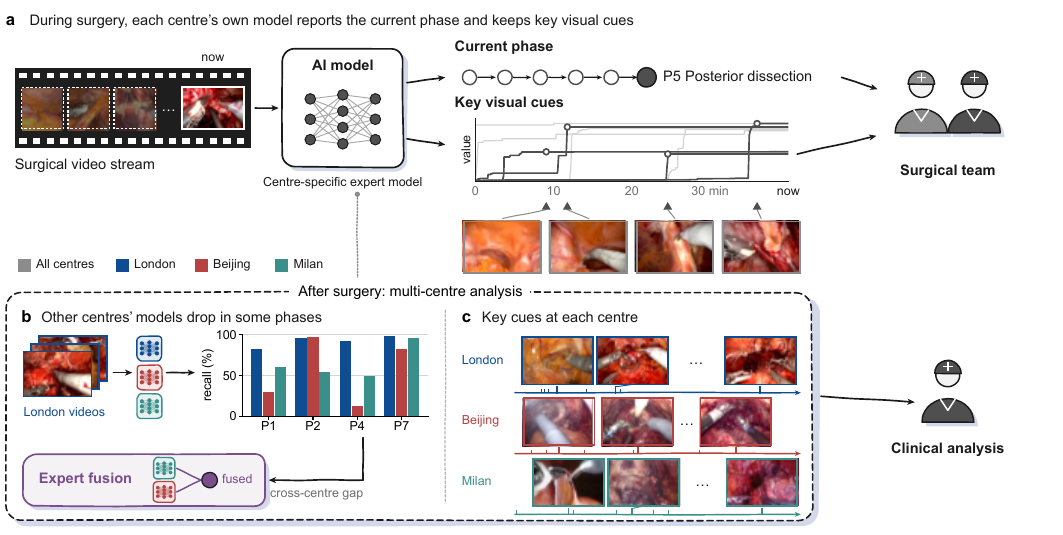}
\caption{Study overview. (a) During surgery, a centre-specific expert model reads the surgical video stream, updates a fixed-size key memory and outputs the current phase. (b) After surgery, cross-centre testing on London recordings and expert fusion by probability averaging. (c) Key visual cues traced from memory at each centre.}
\label{fig:overview}
\end{figure}

In this study, we further investigate the ability of SKiT, our streaming AI model incorporating temporal memory, to accurately recognise and analyse surgical workflows across three international centres performing robot-assisted radical prostatectomy (RARP) (Fig.~\ref{fig:overview}).
\section*{Results}
\subsection*{Three-centre cohorts and a streaming model with bounded memory}
The study comprised 114 surgical recordings: 39 from London, 40 from Milan and 35 from Beijing (Table~\ref{tab:cohort}). For ethical reasons, recordings from different centres were not pooled, so we trained one model per centre. Each model was evaluated on its own centre's held-out recordings, which gives a local reference. The models trained in Milan and Beijing (hereafter the Milan and Beijing experts) were also applied to the same nine London recordings. Neither centre's temporal model was trained or selected on London recordings, so the London reference gives the in-centre level for this target.

\begin{figure}[htbp]
\centering\includegraphics[width=\linewidth]{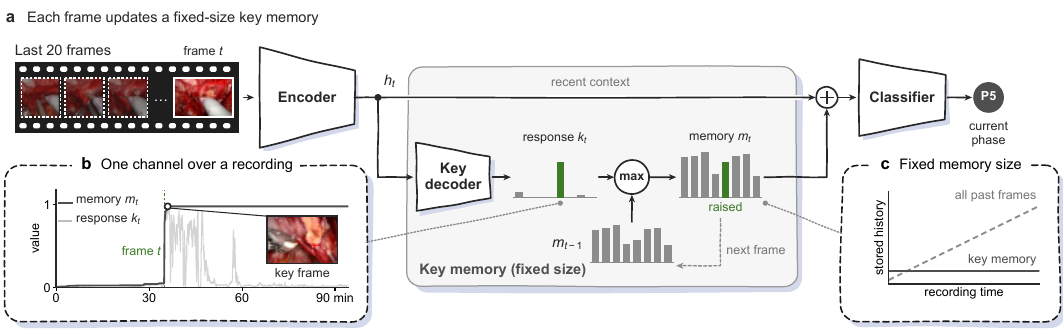}
\caption{Streaming model with a fixed-size key memory. (a) Each new frame updates the key memory by a running maximum; the memory and the recent context together give the current phase. (b) One memory channel over a recording and the frame at which it reached its final maximum (compact London model, recording L03). (c) Memory size does not grow with recording length.}
\label{fig:framework}
\end{figure}

SKiT combines the most recent 20 frames (20~s at 1~Hz) with a fixed-size global key memory (Fig.~\ref{fig:framework}). Each memory channel retains its largest response observed so far, so earlier visual context can contribute without storing an expanding sequence of features. Recognition and duration analyses used the primary models, each with 64 memory channels. To make the memory inspectable, we also trained compact models with eight memory channels at each centre; these were used only for the visual analysis in Fig.~\ref{fig:memory-scenes}. Image-to-output inference took a mean of 10.49~ms per frame on a graphics processing unit (GPU) and 320.67~ms on four central processing unit (CPU) threads, both within the 1-s sampling interval.
\begin{table}[htbp]\centering\small
\caption{Cohorts, partitions and analysis roles. The London evaluation count excludes one duplicate recording; phase codes are defined in Supplementary Table~\ref{stab:taxonomy}.}
\label{tab:cohort}
\begin{tabular}{lrrll}\toprule
\multirow{2}{*}{Centre} & \multicolumn{2}{c}{Recordings ($n$)} & \multirow{2}{*}{Label schema} & \multirow{2}{*}{Role}\\
\cmidrule(lr){2-3}
 & Training & Evaluation & & \\\midrule
London & 30 & 9 & P0--P7, P9 & Common target; local reference\\
Milan & 30 & 10 & P0--P9 & Source expert; local reference\\
Beijing & 28 & 7 & P0--P10 & Source expert; local reference\\\bottomrule
\end{tabular}\end{table}

\subsection*{Annotated workflows share a progression but differ in phase order}
\begin{figure}[!htb]
\centering\includegraphics[width=\linewidth]{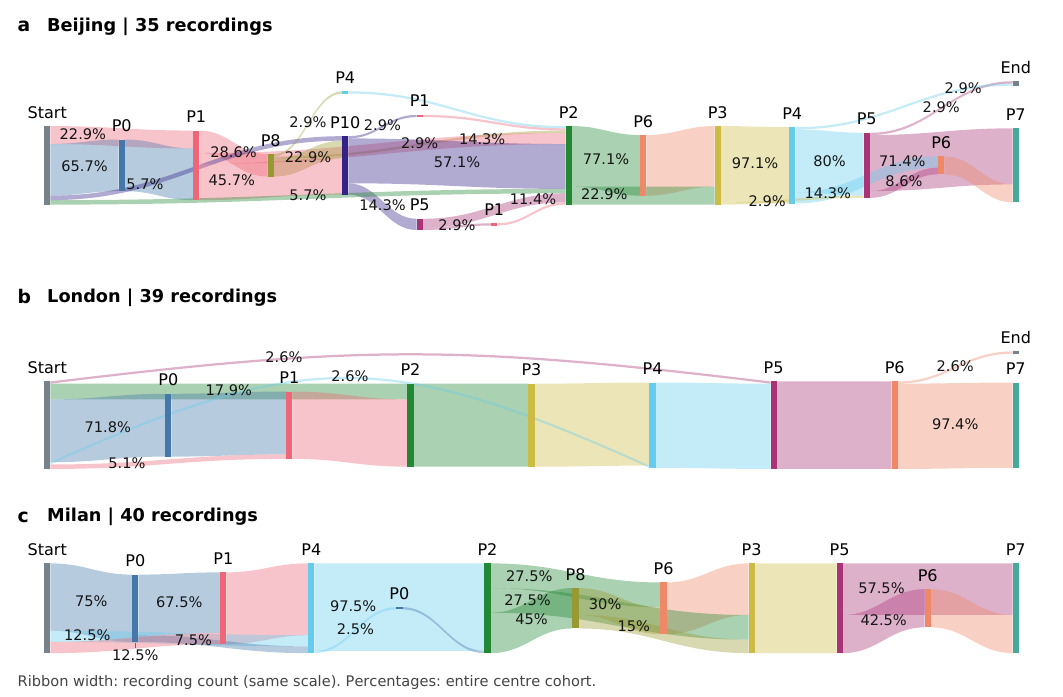}
\caption{First-appearance order in all unique recordings: Beijing ($n=35$), London ($n=39$) and Milan ($n=40$). Ribbon widths encode recording counts; percentages use each full cohort as denominator. Sequences retain each phase's first occurrence up to the first P7 or recording end, omitting P9 and unlabelled footage. Links encode order, not direct transitions or completion. A phase appears as more than one node when its position differs between recordings, and End denotes recordings without P7. Definitions appear in Supplementary Table~\ref{stab:taxonomy}.}
\label{fig:gt-workflows}
\end{figure}

Annotations showed a common overall progression but centre-specific differences in first-appearance order (Fig.~\ref{fig:gt-workflows}). All 39 London recordings followed a single first-appearance order, whereas 25 of 35 Beijing and 22 of 40 Milan recordings followed their centre's majority order. Apical dissection and dorsal venous complex control (P6) preceded anterior bladder-neck transection (P3) in 27 of 35 Beijing recordings and in no London recording. In Milan, posterior bladder-neck and seminal-vesicle dissection (P4) preceded bladder mobilisation (P2) in all 40 recordings and was labelled again after other phases in 39 of them. Lymphadenectomy (P8) was labelled in 18 Milan recordings, always before P3, and P6 preceded P3 in 23.

The evaluation sets showed the same difference: P6 preceded P3 in six of seven Beijing recordings and in no London recording. The local Retzius-approach label (P10) occurred in five Beijing evaluation recordings. Annotation timelines, relative phase durations and re-entry summaries are shown in Supplementary Figs.~\ref{sfig:gt}, \ref{sfig:durations} and \ref{sfig:workflow}.

\subsection*{Expert rankings reverse between surgical phases}
As shown in Table~\ref{tab:performance}, both experts were less accurate on the London target than at home. On this target, the Milan model achieved 72.27\% frame accuracy (95\% whole-recording bootstrap interval, 64.25--78.50) and the Beijing model 62.66\% (55.63--69.05). Their paired difference was $-9.61$ percentage points ($-19.49$ to 1.19). Phase-average recall was 68.54\% and 56.52\%, respectively. For comparison, the local references reached 92.95\% in London, 87.14\% in Milan and 89.35\% in Beijing.
\begin{table}[htbp]\centering\small
\caption{Recognition and phase-duration results. Frame, Phase and Video are percentages: pooled frame accuracy, equally weighted phase recall and equally weighted recording accuracy, respectively. Brackets give 95\% whole-recording bootstrap confidence intervals (CI). MAE is mean absolute phase-duration error in minutes. The experts and probability mean use the same nine London recordings; local references use their own centre's evaluation recordings.}
\label{tab:performance}
\begin{tabular}{lrrrrr}
\toprule
Model / rule & $n$ & Frame (\%) [95\% CI] & Phase (\%) & Video (\%) & MAE (min)\\\midrule
\multicolumn{6}{l}{\textit{Local references}}\\
London reference & 9 & 92.95 [89.12, 95.66] & 90.29 & 92.87 & 1.30\\
Milan reference & 10 & 87.14 [83.51, 90.00] & 81.07 & 87.38 & 1.85\\
Beijing reference & 7 & 89.35 [78.28, 95.55] & 89.31 & 88.48 & 1.14\\
\addlinespace[0.5em]
\multicolumn{6}{l}{\textit{Experts and probability mean on London}}\\
Milan & 9 & 72.27 [64.25, 78.50] & 68.54 & 71.15 & 4.58\\
Beijing & 9 & 62.66 [55.63, 69.05] & 56.52 & 62.11 & 6.83\\
Probability mean & 9 & 75.43 [69.93, 79.96] & 70.00 & 75.18 & 4.44\\
\bottomrule\end{tabular}

\end{table}

\begin{figure}[htbp]
\centering\includegraphics[width=\linewidth]{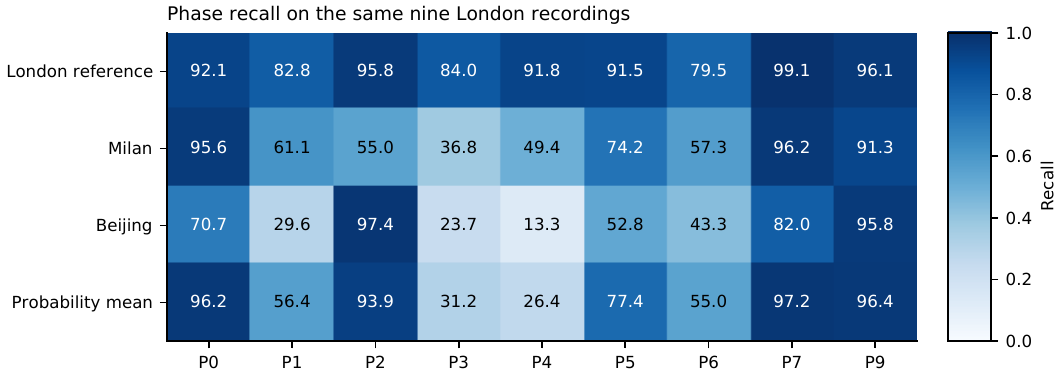}
\caption{Phase recall of the London reference, the Milan and Beijing experts and their probability mean (equal probability averaging) on the same nine London recordings. Cells give recall in per cent. P8 and P10 lie outside the London schema; predictions of these labels count as errors. Phase codes are defined in Table~\ref{tab:cohort}.}
\label{fig:transfer}
\end{figure}

The two models had differing phase-level strengths (Fig.~\ref{fig:transfer}). Beijing had higher recall for bladder mobilisation (P2), 97.39\% against 54.98\% for Milan, whereas Milan had higher recall for posterior bladder-neck dissection (P4), 49.44\% against 13.25\%. Both models recognised anterior bladder-neck transection (P3) poorly (36.76\% and 23.68\%). Differences persisted in the F1 score, which also penalises false positives. For P2, F1 was 76.41\% for Beijing and 68.99\% for Milan; for P4, it was 64.08\% for Milan and 23.40\% for Beijing (Supplementary Tables~\ref{stab:phase} and \ref{stab:extended}). The reversal did not extend to duration error, which was lower for Milan in both phases (P2, 6.80 versus 9.40 minutes; P4, 7.28 versus 13.81 minutes).

Confusion matrices identified the substitutions behind these deficits. Milan assigned 28.67\% of P4 frames to posterior and neurovascular-bundle dissection (P5), whereas Beijing assigned 48.08\% of P4 frames to P5 and 73.16\% of P3 frames to P2 (Supplementary Fig.~\ref{sfig:confusion}).

The two experts failed in different ways. Beijing recalled most P2 frames but over-assigned P2 (precision 62.87\%), absorbing most P3 frames. Milan missed P2 more often but rarely over-assigned it (precision 92.56\%). P2 and P4 are also the phases whose annotated order differed in Milan, where P4 preceded P2 (Fig.~\ref{fig:gt-workflows}).

\subsection*{Transfer degrades phase-duration estimates}
\begin{figure}[!t]
\centering\includegraphics[width=\linewidth]{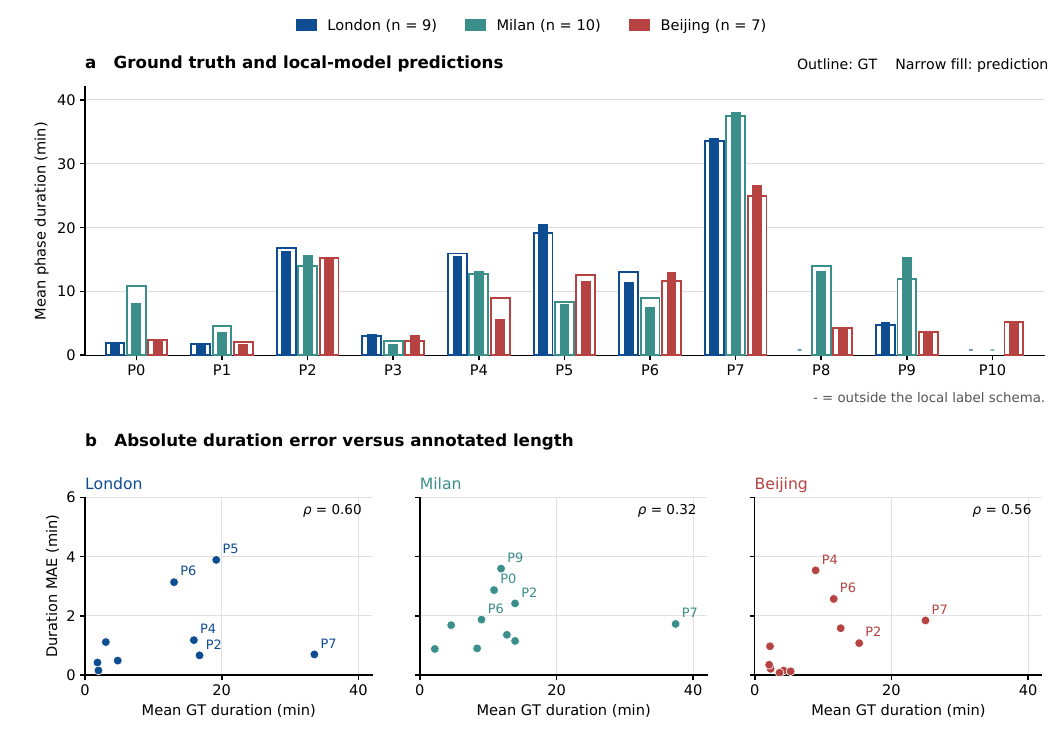}
\caption{Within-centre phase-duration assessment in London ($n=9$), Milan ($n=10$) and Beijing ($n=7$). (a) Each centre--phase position shows mean annotated (GT) duration as an outline and mean predicted duration as a narrower filled bar. Dashes mark labels outside the local schema. (b) Phase-level mean absolute error versus mean GT duration; $\rho$ is descriptive Spearman correlation, with selected phases labelled. Errors are computed per recording before averaging. Surgical techniques differ between centres, so durations should not be compared across centres.}
\label{fig:duration}
\end{figure}

We next asked whether recognised phases yield usable durations. Within their own centres, the three models estimated phase durations with mean absolute errors of 1.30 (London), 1.85 (Milan) and 1.14 minutes (Beijing). Mean predicted durations followed annotated durations (Fig.~\ref{fig:duration}a), and the relation between phase length and error varied between centres (Fig.~\ref{fig:duration}b). On the nine London recordings, the Milan and Beijing experts reached errors of 4.58 and 6.83 minutes (Table~\ref{tab:performance}). The largest transfer errors were Beijing's 13.81 minutes in posterior bladder-neck dissection (P4) and 12.96 minutes in vesicourethral anastomosis (P7; Supplementary Table~\ref{stab:extended}). For P4, Beijing assigned 2.12 minutes per recording against 15.93 annotated minutes.

After transfer, high recall did not guarantee reliable phase duration. Beijing recognised 97.39\% of bladder-mobilisation (P2) frames but assigned 25.96 minutes per recording to P2 against 16.76 annotated minutes, a mean absolute error of 9.40 minutes. Milan recognised 91.33\% of extracorporeal (P9) frames but assigned 14.84 minutes against 4.79 annotated minutes, a mean absolute error of 10.05 minutes. In both cases, false-positive intervals inflated the estimated duration without reducing recall.

Segment-level agreement was also lower after transfer (Supplementary Tables~\ref{stab:segment} and \ref{stab:boundary}). The segmental Edit score was 29.85 for the London reference and 4.57 and 6.77 for the Milan and Beijing experts.

\subsection*{Probability averaging trades errors between phases}
Because the two experts performed better in different phases, we tested whether equal probability averaging would combine their strengths. It reached 75.43\% frame accuracy, 3.16 percentage points above Milan (paired 95\% interval, $-1.25$ to 9.15). Mean absolute duration error changed from 4.58 to 4.44 minutes (paired difference, $-0.14$ minutes; $-1.00$ to 0.65).

Phase-level results explain why the aggregate change was small. Averaging raised recall for bladder mobilisation (P2) from 54.98\% to 93.89\% and reduced its duration error from 6.80 to 3.73 minutes. For posterior bladder-neck dissection (P4), recall fell from 49.44\% to 26.36\% and duration error rose from 7.28 to 11.59 minutes. Anastomosis (P7) recall was 97.15\% (Milan, 96.19\%), but its duration error rose from 2.70 to 6.32 minutes. Averaging improved one part of the timeline at the expense of another, so its value depends on the phase under review (Supplementary Fig.~\ref{fig:cases}; Supplementary Tables~\ref{stab:fusion} and \ref{stab:subsets}).

\subsection*{Compact-model memory maxima recur within particular phases}
\begin{figure}[p]
\centering\includegraphics[width=\linewidth]{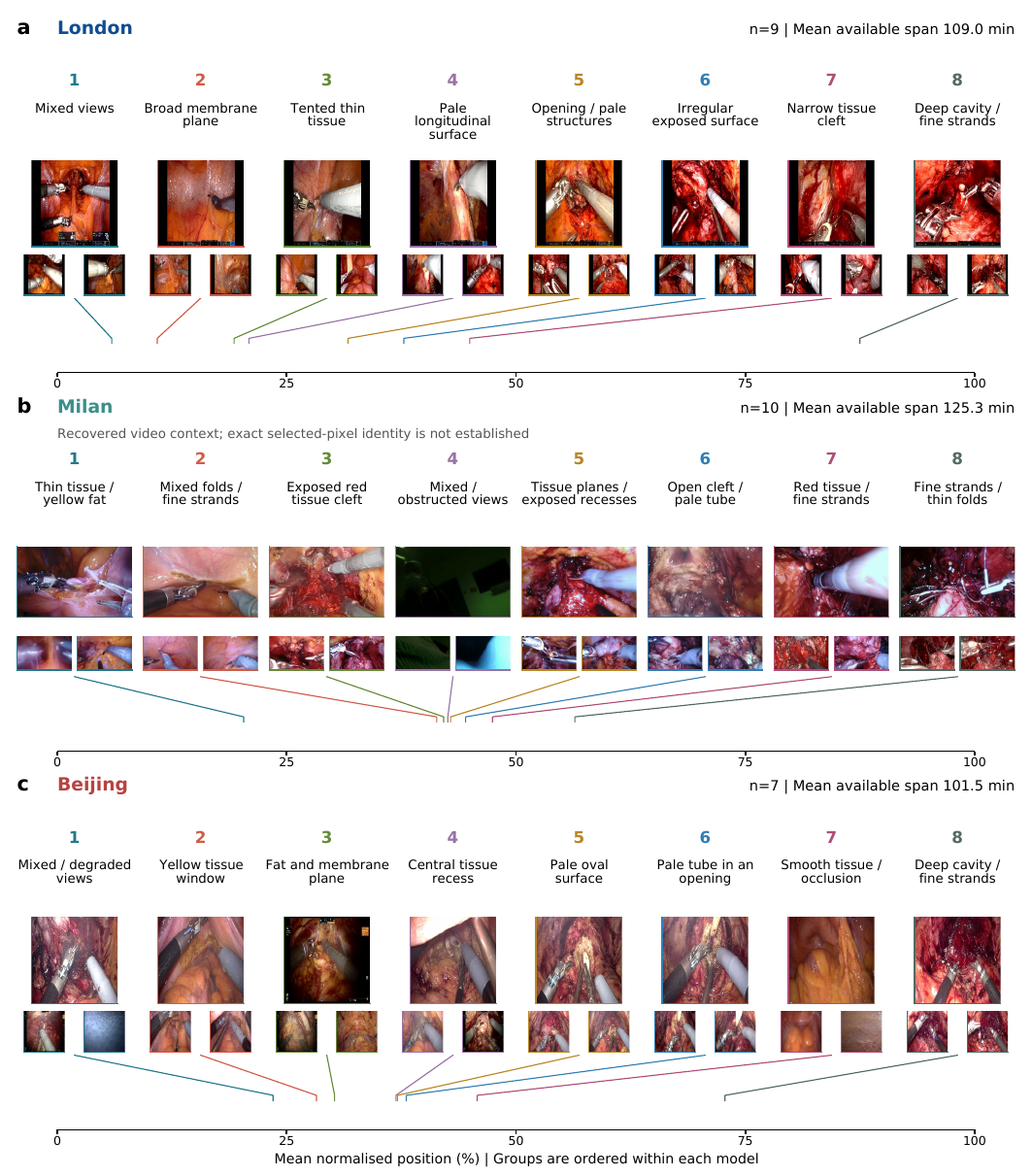}
\caption{Visual contexts associated with compact-model memory in London (a, $n=9$), Milan (b, $n=10$) and Beijing (c, $n=7$). Connectors indicate mean selection times normalised within recordings. Groups are ordered separately within each model. The large image is the selected frame most typical of the group, and the two small images are the next most typical. Milan images are the frames nearest to each selected sample. Descriptions are exploratory. Supplementary Tables~\ref{stab:timing} and \ref{stab:groups} describe the London and Beijing models.}
\label{fig:memory-scenes}
\end{figure}

Because retained memory affected recognition, we next examined what a memory retained. For each compact model, we traced every memory channel to the video frame at which it first reached its final value (Fig.~\ref{fig:memory-scenes}). This gave 208 selected frames across nine London, ten Milan and seven Beijing recordings. Channels were learned independently at each centre, so they are described within centres. The frames selected by one channel form a group, which we name by the visible scene in its most typical frames.

In London, the broad-membrane group selected bladder mobilisation (P2) in eight of nine recordings and the tented-tissue group selected adhesiolysis (P1) in seven. The late deep-cavity group selected vesicourethral anastomosis (P7) in all eight recordings with a P7 annotation. Two further groups were each concentrated in one phase. The pale-surface group selected anterior bladder-neck transection (P3) in all nine recordings, and the narrow-cleft group selected posterior and neurovascular-bundle dissection (P5) in eight. The remaining three groups had mixed labels (Supplementary Table~\ref{stab:groups}).

In Beijing, the yellow-window group selected the local Retzius-approach label (P10) in five of seven recordings, and the late deep-cavity group selected P7 in all seven. The oval-surface and tube-in-opening groups selected anterior bladder-neck transection (P3) and posterior bladder-neck dissection (P4) in six of seven recordings each, and neither selected apical dissection (P6). These two groups therefore point clinical review to bladder-neck rather than apical contexts.

In Milan, groups~1 and 5 selected P2 in six and seven of ten recordings, and group~6 selected P4 in nine. Group~7 selected posterior and neurovascular-bundle dissection (P5) in eight recordings and lymphadenectomy (P8) in two.
The latest group selected P7 in seven recordings but P1
in three.

The order in which the eight channels reached their maxima was consistent in London but varied between recordings in Milan and Beijing. Concordance of channel order (Kendall's $W$) was 0.665 in London but 0.181 in Milan and 0.252 in Beijing, only modestly above the 0.10--0.14 expected under random ordering (Supplementary Fig.~\ref{sfig:channel-order}). This ranking matched the annotations, in which phase order was uniform in London and most variable in Milan (Fig.~\ref{fig:gt-workflows}). In Milan, concentrated P4 and P5 associations coexisted with low order concordance. Timing, label counts, and candidate interpretation checks are given in Supplementary Note~4 and Tables~\ref{stab:timing}--\ref{stab:hypotheses}.

\clearpage
\section*{Discussion}

This multicentre retrospective feasibility study evaluated the ability of a streaming AI model with traceable key memory (SKiT) to recognise and analyse surgical phases during robot-assisted radical prostatectomy (RARP). The model demonstrated high within-centre performance across three distinct institutions, with video-level accuracy ranging from 87\% to 93\%. These findings are particularly encouraging given the geographical diversity of the participating centres and differences in surgical techniques, operative approaches and environments. The model also demonstrated computational efficiency, with an inference time of 10.49~ms per frame on a GPU and a fixed memory state of 60.25~KiB, independent of recording length, supporting its potential application for real-time intraoperative phase recognition. However, when models were transferred between centres, recognition accuracy decreased substantially, with marked differences in performance across individual surgical phases. Importantly, high phase recall did not necessarily translate into accurate estimates of phase duration, highlighting the challenges of generalising AI-derived surgical workflow measurements across institutions. 

Analysis of the model's traceable key memory identified reproducible associations between retained visual information and specific surgical phases, providing a potential approach to improving the explainability of surgical AI. Each memory channel retains its maximum response, allowing the historical video frame responsible for that response to be identified directly without requiring additional post-hoc attribution methods. Despite being trained only on surgical phase labels, individual channels in compact eight-channel models consistently selected frames associated with particular operative phases. For example, one channel in the London model selected anterior bladder-neck transection (P3) in all nine recordings, while channels in the Beijing and Milan models selected the local Retzius-approach phase (P10) in all five recordings containing that phase and posterior bladder-neck dissection (P4) in nine of ten recordings, respectively. Furthermore, the temporal order in which memory channels reached their maximum responses was more consistent in London than in Milan or Beijing, broadly reflecting differences in the consistency of surgical phase sequences between centres. These findings suggest that the model retains information associated with operative context and progression, providing an opportunity to examine the historical visual information retained during surgical phase recognition. This traceability may facilitate clinical interpretation and validation of model behaviour, although memory ordering does not represent surgical skill, and the selected frames do not establish which anatomical features influence individual predictions. Further independent clinical annotation and targeted interventions will be required to establish the anatomical significance and functional contribution of individual memory channels \cite{bau2017,kim2018}.

Previous studies have demonstrated the feasibility of automated surgical phase recognition in RARP, including validation across different surgeons and robotic platforms \cite{konnai2026,ueki2025}. Konnai et al. reported a reduction in recognition performance when models were applied to recordings from different surgeons, while Ueki et al. demonstrated substantial deterioration in accuracy when transferring between robotic platforms. Zhao et al. further demonstrated the clinical potential of phase recognition in a multicentre study involving 18 institutions, using AI-derived phase durations and other measures to differentiate surgical expertise \cite{zhao2025rarp}. Similar challenges in cross-centre generalisation have been reported in other surgical procedures. Lavanchy et al. demonstrated substantial performance deterioration when transferring phase-recognition models between gastric bypass centres, with improved generalisation following multicentre training \cite{lavanchy2024}. In contrast, Sui et al. reported relatively modest reductions in surgical-step recognition when evaluating models trained on multicentre cholecystectomy datasets against previously unseen institutions \cite{sui2026}. Our study demonstrates that cross-centre generalisation errors are not uniform across surgical phases, with models trained at different institutions showing distinct phase-specific strengths when evaluated on identical recordings. Importantly, we also demonstrate that high phase-recognition recall does not necessarily translate into accurate estimates of surgical phase duration. These findings highlight the need to evaluate the reliability of AI-derived surgical workflow measurements, particularly when models are transferred between institutions, rather than relying solely on overall recognition accuracy.

This study has also shown that, for workflow measurement, false positives are arguably as important as missed phases. They inflate estimated duration despite high recall, whereas missed and added intervals can cancel in a duration total despite poor temporal localisation. When recognised phases are used to construct surgical performance measures such differences can have important repercussions\cite{zhao2025rarp,reinke2024}. A recent review of validation practice in robotic surgical guidance similarly calls for clinically relevant, transparently reported evaluation metrics\cite{korsten2026}.

This study has several limitations. First, cross-centre generalisation was evaluated on a relatively small cohort of nine London recordings, limiting the generalisability of these findings to other institutions and surgical settings. Although a common surgical phase definition protocol was used across all three centres, differences in surgical techniques, training datasets and visual encoders mean that the factors contributing to differences in model performance cannot be established independently. Second, each model was trained only once, and the evaluation recordings had been used in earlier analyses. Consequently, the bootstrap confidence intervals reflect uncertainty associated with the sampled recordings rather than variability in model training or repeated evaluation. Third, inter-annotator agreement and relevant clinical characteristics, including case complexity and surgeon experience, were not assessed. Fourth, probability averaging was exploratory, with equal weighting of the models and no parameters optimised using London annotations. 

In summary, this multicentre study demonstrates the feasibility of accurate and computationally efficient streaming analysis of robotic surgical video, highlighting its potential to support real-time intraoperative feedback, contextual guidance and surgical planning. Further improvements in temporal stability\cite{liu2026stablespr} and cross-centre generalisability will be essential for clinical translation, representing an important step towards the integration of intelligent, context-aware assistance into robotic surgery.
\section*{Methods}
\subsection*{Ethics and data governance}
The study followed the Declaration of Helsinki. Use of the London and Beijing recordings was approved by the London--Surrey Research Ethics Committee (REC 23/LO/0492; IRAS 312298) and the Biomedical Research Ethics Committee of Peking University First Hospital (2025R0457-0002). The Milan study was notified to the Comitato Etico Territoriale Lombardia 1 as an observational study (CET 189-2024), and its patients had consented within earlier institutional studies.

\subsection*{Study design, partitions and annotations}
This retrospective study used 114 RARP recordings from three centres listed in the author affiliations. Training/evaluation partitions were 30/9 for London, 30/10 for Milan and 28/7 for Beijing, and no recording was used for both. The evaluation recordings had a mean duration of 109.0 minutes in London, 125.3 minutes in Milan and 101.5 minutes in Beijing and were sampled at a nominal 1~Hz. Recordings were the resampling units. Input details appear in Supplementary Note~1.

A common surgical phase definition protocol was used across all centres, although the phases represented in each dataset differed according to the surgical activities performed: London (P0--P7 and P9), Milan (P0--P9) and Beijing (P0--P10)(Supplementary Table~\ref{stab:taxonomy}). For example, pelvic lymphadenectomy (P8) was not performed in the London cohort. Missing phase labels reflected differences in the surgical activities performed and the annotation categories used, rather than inconsistencies in phase definitions\cite{meireles2021,mlambo2025}.

\subsection*{Annotation-based workflow reference}
The full-cohort order diagram combines training and evaluation annotations and is descriptive. Each sequence was truncated at the first P7 or recording end; P9 and unlabelled samples were omitted, and only each remaining phase's first occurrence was retained. Milan interval annotations were expanded to nominal 1~Hz, mapping extracorporeal intervals to P9. A recording was counted as following its centre's majority order when every pair of its phases first appeared in the order seen in most of that centre's recordings. Evaluation timelines retain unlabelled intervals. Separate re-entry and adjacent-transition summaries preserve gaps rather than bridging them (Supplementary Note~2).

\subsection*{Streaming model and training}
The framework uses the previously reported SKiT~\cite{liu2023skit} architecture without introducing a new network module. SKiT combines recent visual context with a fixed-size key memory (Fig.~\ref{fig:framework}); memory channels are learned through phase supervision without clinical key-event annotations. We call a model online when
its prediction at time $t$ uses no later frames, and streaming when its memory and per-frame computation do not grow with video length. SKiT is both.

SKiT predicts phase $y_t$ from observations available up to time $t$ in a sampled video $\{x_1,\ldots,x_T\}$. Each source-specific visual encoder produced a feature vector $\mathbf f_t$, which a linear projection $\mathbf W_{\mathrm{in}}$ mapped to the model dimension. The local transformer received a window of $W=20$ samples ending at $t$, with zero padding before the start of the recording and learned positional embeddings within the window. The representation of the current sample was the last output token:
\begin{equation}
\mathbf h_t =
\left[\operatorname{Transformer}\left(
\{\,\mathbf W_{\mathrm{in}}\mathbf f_i+\mathbf e_{i-t+W}\,\}_{i=t-W+1}^{t}
\right)\right]_{W}.
\end{equation}
Tokens attended only within the window of current and past samples, so the prediction at $t$ had no access to later frames.

A learned decoder compresses $\mathbf h_t$ into a $d$-dimensional key:
\begin{equation}
\mathbf k_t=\operatorname{softmax}
\left(\operatorname{ReLU}(\mathbf W_k\mathbf h_t+\mathbf b_k)\right).
\end{equation}
The key memory updates each channel independently:
\begin{equation}
m_{t,j}=\max(m_{t-1,j},k_{t,j})
       =\max_{1\leq i\leq t} k_{i,j},\qquad j=1,\ldots,d,
\quad \mathbf m_0=\mathbf 0.
\label{eq:recorder}
\end{equation}
Because key values are non-negative, the memory can start at zero.

Equation~\ref{eq:recorder} retains the largest observed response in each channel, and different channels can reach their maxima at different times. The memory stores neither frame identities nor the full history of tokens. Each $\mathbf k_t$ is softmax-normalised, but the accumulated vector $\mathbf m_t$ need not sum to one and does not represent phase probabilities.

The memory preserves response magnitude but not occurrence counts or explicit event order; recent context remains available through the local branch. Because each key response depends on a local window, interpreting its maximum requires examining the preceding video context as well as the selected frame.

Memory storage and each update cost $O(d)$ and do not grow with recording length. Local attention uses the fixed window $W$, so its input size also remains bounded as the procedure progresses. These bounds apply to the temporal model; visual features are computed separately.

The memory state is mapped back to the local feature dimension by a linear layer, ReLU and layer normalisation:
\begin{equation}
\mathbf g_t=\operatorname{LN}\!\left(
\operatorname{ReLU}(\mathbf W_m\mathbf m_t+\mathbf b_m)\right).
\end{equation}
The resulting memory representation is added to the local feature, followed by one linear residual block and a classification layer:
\begin{equation}
\mathbf u_t=\mathbf h_t+\mathbf g_t,\qquad
\mathbf r_t=\mathbf u_t+\mathbf W_r\mathbf u_t,\qquad
\mathbf z_t=\mathbf W_c\mathbf r_t+\mathbf b_c .
\end{equation}
All primary models produced eleven class logits in a common label order. During training, an auxiliary head predicted phase transitions; inference used only the class logits.

Each model's memory was reset at the start of every recording and carried forward through it, with no post-hoc smoothing. Each primary model was trained once on its centre's training partition and used without selection on evaluation recordings; training settings are given in Supplementary Note~1.

\subsection*{Recognition and temporal-localisation endpoints}
For a target schema $S_T$ and annotated ground-truth (GT) label $y_t$, the valid-label mask was $M_t=\mathbf{1}\{y_t\in S_T\}$. Frame accuracy pooled correct predictions over valid samples. With true-positive (TP), false-positive (FP) and false-negative (FN) samples counted per phase, recall was $\mathrm{TP}/(\mathrm{TP}+\mathrm{FN})$; phase-average recall weighted supported phases equally, and video accuracy weighted recordings equally. All eleven predicted labels remained possible, so unsupported target labels counted as errors. Precision and F1 included false positives over all valid samples. Full phase metrics and GT-subset comparisons appear in Supplementary Tables~\ref{stab:extended} and \ref{stab:subsets}.

Unsmoothed label runs formed segments. The Edit score was $100(1-e/n)$, where $e$ is the Levenshtein distance between the predicted and annotated run-label sequences and $n$ is the length of the longer sequence. Higher values indicate better agreement, and scores were averaged equally over recordings. Segment F1 pooled one-to-one same-phase matches at intersection-over-union thresholds of 10\%, 25\% and 50\% (F1@10, F1@25 and F1@50), following action-segmentation definitions\cite{farha2019mstcn}. P9 was retained and unsupported predictions counted as errors. Boundary analysis matched identical ordered phase transitions one-to-one by minimum absolute sample-index difference, reporting matched and total counts together.

\subsection*{Phase-duration endpoints}
For recording $v$ and phase $p$, annotated and predicted durations in minutes were
\begin{equation}
D_{vp}=\frac{1}{60}\sum_{t\in v}M_t\mathbf{1}\{y_t=p\},\qquad
\widehat D_{vp}=\frac{1}{60}\sum_{t\in v}M_t\mathbf{1}\{\hat y_t=p\}.
\end{equation}
Absolute error was computed within each recording before averaging: $\mathrm{MAE}_p=V^{-1}\sum_v|\widehat D_{vp}-D_{vp}|$. The overall MAE averaged these phase-specific errors equally over the supported schema. Every recording contributed, with zero duration for an absent supported phase. At 1~Hz, signed error equals $(\mathrm{FP}-\mathrm{FN})/60$; missed and added intervals can cancel, so duration agreement does not establish overlap. Durations were computed from labelled samples rather than operative wall-clock time. Local reference plots and the Milan repeated-endpoint sensitivity are detailed in Supplementary Notes~1 and 2.

\subsection*{Exploratory probability averaging}
Each expert's eleven-class logits were converted to probabilities at temperature one. Equal averaging used $\mathbf p_t^{\mathrm{mean}}=(\mathbf p_t^{M}+\mathbf p_t^{B})/2$, followed by arg\,max; probability assigned to P8/P10 remained included. No routing, temperature or phase-specific parameter was fitted to London labels, and no temporal post-processing was applied. The decision to study averaging followed inspection of the target results, so the analysis remains exploratory. Fixed confidence-gate and entropy-weighted comparators are defined in Supplementary Note~2 and reported in Supplementary Table~\ref{stab:fusion}.

\subsection*{Eight-channel visual contexts and temporal concordance}
The visual analysis used compact eight-channel models trained separately in London, Milan and Beijing on the same training partitions as the primary models. Each was trained once, without model selection on evaluation recordings, and evaluated on its own centre's evaluation recordings.

For each recording and channel, we retrospectively selected the first sample attaining the final cumulative maximum. This analysis requires the completed sequence, whereas online phase prediction uses only the running state available at the current sample. One London trajectory was constant, so its displayed time of zero denotes an initial record rather than a detected event. 

Duplicate images were counted once within each channel group when choosing a representative. Selected spatial features were L2-normalised, and the selected sample nearest their mean provided the representative illustration. The next two nearest unique samples were displayed at smaller sizes. Descriptions were limited to visible scene properties, and channels from independently trained models were not aligned across centres.

Timing summaries weighted the unique recordings equally. Within each recording, the eight selection times were ranked, with average ranks for ties. Tie-corrected Kendall's $W$ measured agreement in channel order across recordings\cite{kendall1939}: $W=1$ denotes identical ordering, and lower values denote less agreement. The measure is unchanged by consistently renaming channels within a model or by positively rescaling times within a recording. Leave-one-recording-out ranges were sensitivity checks, not uncertainty intervals.

With timestamps starting at zero, relative position was $\overline{\tau}_j=(100/N)\sum_{v=1}^{N}t_{vj}/T_v$, where $T_v$ is the final available timestamp of recording $v$; normalisation thus preceded averaging. Figure headers report mean $T_v$ in minutes. P8/P10 counts describe labels at selected positions without adjustment for phase exposure. Retention spans ran from each selected sample to the final input sample.

Candidate interpretations combined visible scene properties, local composite phase definitions and published RARP technique descriptions (Supplementary Table~\ref{stab:hypotheses}). They define questions for clinical review and were not independently annotated by clinicians.

Each of the 208 selections was linked to the GT label at its source position.

\subsection*{Statistical analysis}
Recognition and duration intervals are 95\% percentile bootstrap intervals from 10,000 resamples of complete recordings. Confusion-count metrics were recomputed per resample, omitting phases without GT support. Video and duration metrics retained equal recording weights. Paired London contrasts used identical recording multiplicities across methods and subsets. With seven to ten recordings per evaluation set, the intervals are descriptive, and no confirmatory hypothesis test was performed.

\subsection*{Inference feasibility}
Latency was measured for the London model by replaying London recordings one frame at a time, from image file to phase probabilities. Measurements used a 40-GB NVIDIA A100 GPU and four CPU threads. The 95th-percentile GPU latency was 12.31~ms, and the state retained between frames occupied 60.25~KiB, excluding model weights. The benchmark excludes live capture, video decoding, queueing, and display. Supplementary Note~3 and Figs.~\ref{fig:inference} and \ref{sfig:inference-jitter} provide the complete measurement scope and distributions.

\begingroup\small\raggedright
\setlength{\bibsep}{1.5pt}
\bibliographystyle{unsrtnat}
\bibliography{references}

\endgroup

\section*{Acknowledgements}
This work was supported by EPSRC (EndoTheranostics/PE/Z003172/1) under the Horizon Europe Guarantee Extension, The Urology Foundation, The John Black Charitable Foundation, and The Recordati Surgical Data Science Programme.

\clearpage
% Supplementary Information (same PDF as the main text; S-numbered floats)
\setcounter{figure}{0}\setcounter{table}{0}\setcounter{equation}{0}
\renewcommand{\thefigure}{S\arabic{figure}}
\renewcommand{\thetable}{S\arabic{table}}
\renewcommand{\theequation}{S\arabic{equation}}
\renewcommand{\theHfigure}{S\arabic{figure}}
\renewcommand{\theHtable}{S\arabic{table}}
\renewcommand{\theHequation}{S\arabic{equation}}
% Supplementary Information body. Included by paper.tex with \input{supplementary}; not a standalone document.
\makeatletter
\setlength{\@fptop}{0pt}
\setlength{\@fpsep}{12pt plus 2pt minus 2pt}
\makeatother
\widowpenalty=10000
\clubpenalty=10000
\begin{center}\Large Supplementary Information\end{center}

\section*{Supplementary Note 1: Data and training details}

\paragraph{Milan annotations.}
Milan annotations are timed intervals, which we expanded to labels at 1~Hz. Consecutive intervals share an endpoint, so some timestamps occurred twice. Evaluation used one prediction per annotated second, which gave 20.8 hours of annotated video in the ten evaluation recordings. Keeping the earlier or the later prediction at a repeated timestamp changed the duration MAE by 0.0003 minutes (1.8453 versus 1.8450).

\paragraph{Training settings.}
Visual features of 768 dimensions were projected to 512 dimensions by a linear layer without bias. The local transformer had four encoder layers, eight attention heads, a 2,048-dimensional feed-forward layer and dropout of 0.2. The auxiliary transition head had a 64-unit hidden layer with ReLU, layer normalisation and a sigmoid output. Each temporal model was trained once for 40 epochs with stochastic gradient descent (learning rate $10^{-4}$, momentum 0.9, weight decay $10^{-5}$), cosine scheduling with warm-up and seed 42. Classification and transition losses were weighted 1 and 2. All results use the model from the final epoch. Supplementary Table~\ref{stab:checkpoint} compares these models with an earlier London training run and an earlier Beijing epoch, which had been selected on the evaluation recordings.

\begin{table}[!htbp]\centering\small
\caption{Phase definitions and the label schema of each centre. P10 keeps the Beijing label `Retzius approach' and does not imply Retzius-sparing surgery.}
\label{stab:taxonomy}
\begin{tabularx}{\linewidth}{ll>{\raggedright\arraybackslash}Xccc}\toprule
ID & Short name & Label definition & London & Milan & Beijing\\\midrule
P0 & Prep & Preparation & Yes & Yes & Yes\\
P1 & Adhes & Adhesiolysis & Yes & Yes & Yes\\
P2 & Drop & Drop bladder from anterior wall and expose apex of prostate & Yes & Yes & Yes\\
P3 & ABN & Anterior bladder neck transection & Yes & Yes & Yes\\
P4 & PBN/SV & Posterior bladder neck and seminal vesicle dissection/retraction & Yes & Yes & Yes\\
P5 & NVB & Posterior dissection and neurovascular bundle control & Yes & Yes & Yes\\
P6 & Apex & Apical dissection of prostate and dorsal venous complex control & Yes & Yes & Yes\\
P7 & Anast & Vesicourethral anastomosis including haemostasis sutures and posterior reconstruction & Yes & Yes & Yes\\
P8 & LND & Lymphadenectomy & No & Yes & Yes\\
P9 & Extra & Extracorporeal footage & Yes & Yes & Yes\\
P10 & Retzius & Retzius approach & No & No & Yes\\\bottomrule
\end{tabularx}
\end{table}

\begin{table}[!htbp]\centering\small
\caption{Primary models and earlier training runs on the nine London recordings. The earlier London run was trained separately and is shown at epochs 40 and 21. The Beijing rows are epochs 40 (the Beijing expert) and 12 of one run. Intervals are 95\% bootstrap intervals over whole recordings.}
\label{stab:checkpoint}
\begin{tabular}{lrrrr}
\toprule
Model & Epoch & Frame (\%) [95\% CI] & Phase recall (\%) & MAE (min)\\\midrule
London reference & 40 & 92.95 [89.12, 95.66] & 90.29 & 1.30\\
\addlinespace[0.5em]
\multirow{2}{*}{Beijing} & 40 & 62.66 [55.63, 69.05] & 56.52 & 6.83\\
 & 12 & 63.46 [56.12, 69.55] & 59.40 & 7.00\\
\addlinespace[0.5em]
\multirow{2}{*}{Earlier London run} & 40 & 87.10 [82.69, 91.02] & 85.10 & 2.48\\
 & 21 & 90.69 [86.32, 94.51] & 88.03 & 1.62\\
\bottomrule\end{tabular}

\end{table}
\clearpage
\section*{Supplementary Note 2: Supporting methods and results}
\subsection*{Annotation summaries}
\begin{figure}[!htbp]
\centering\includegraphics[width=\linewidth]{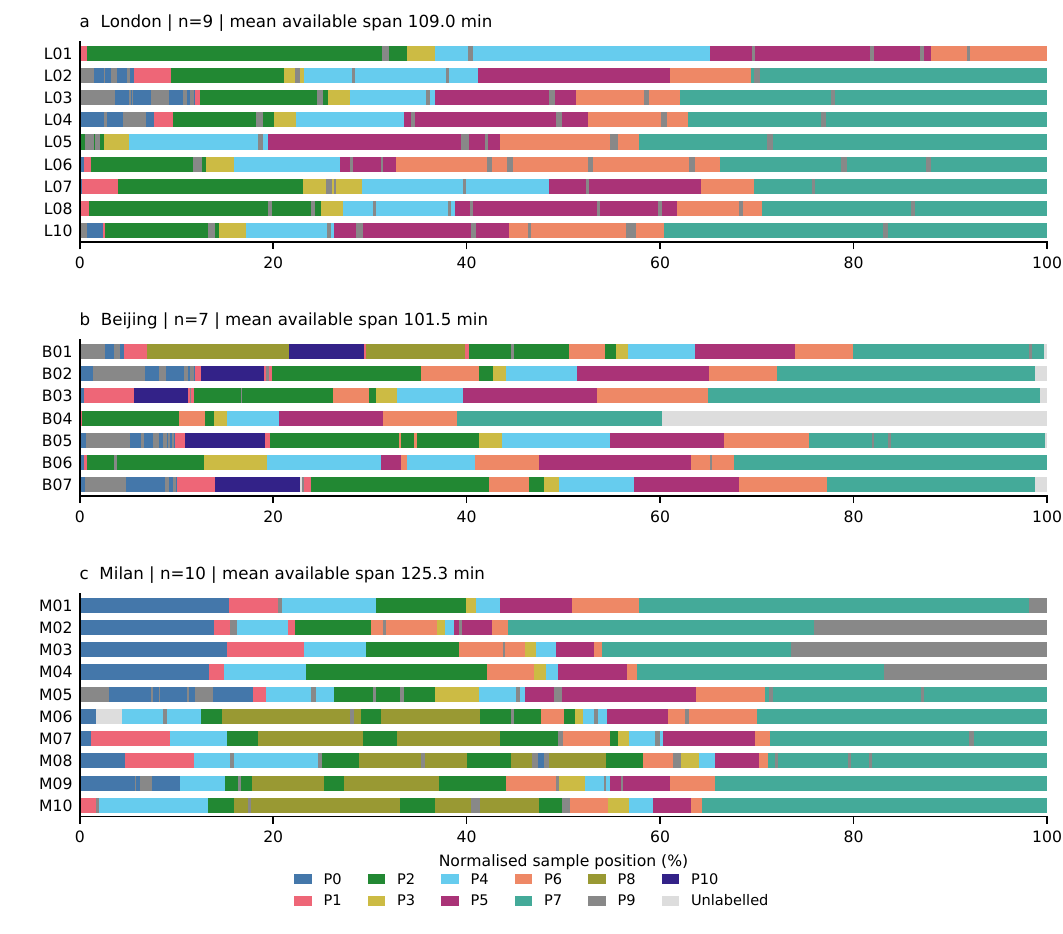}
\caption{Annotated (GT) phase sequences of the evaluation recordings in London ($n=9$), Beijing ($n=7$) and Milan ($n=10$). Light grey marks unlabelled intervals. Position is normalised within each recording, and headers give the mean available span.}
\label{sfig:gt}
\end{figure}

\begin{figure}[!htbp]
\centering\includegraphics[width=\linewidth]{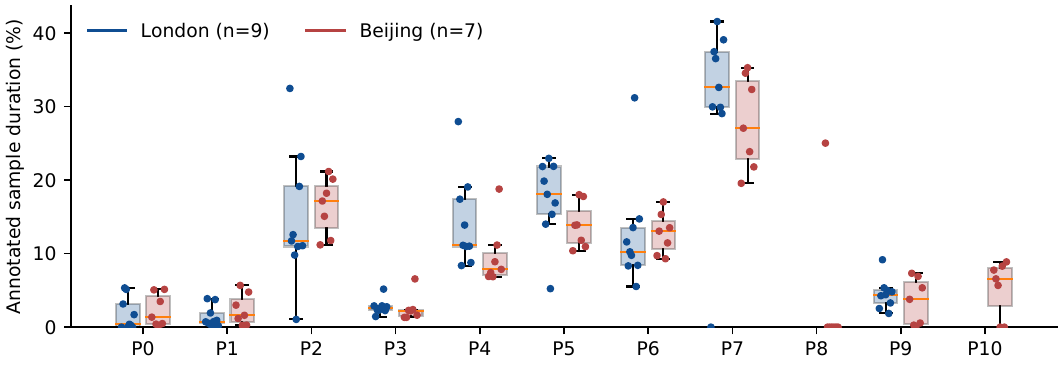}
\caption{Relative annotated phase duration in London ($n=9$) and Beijing ($n=7$). Points are recordings. Boxes show the median and interquartile range, with whiskers to 1.5 times the interquartile range.}
\label{sfig:durations}
\end{figure}

\begin{figure}[!htbp]
\centering\includegraphics[width=\linewidth]{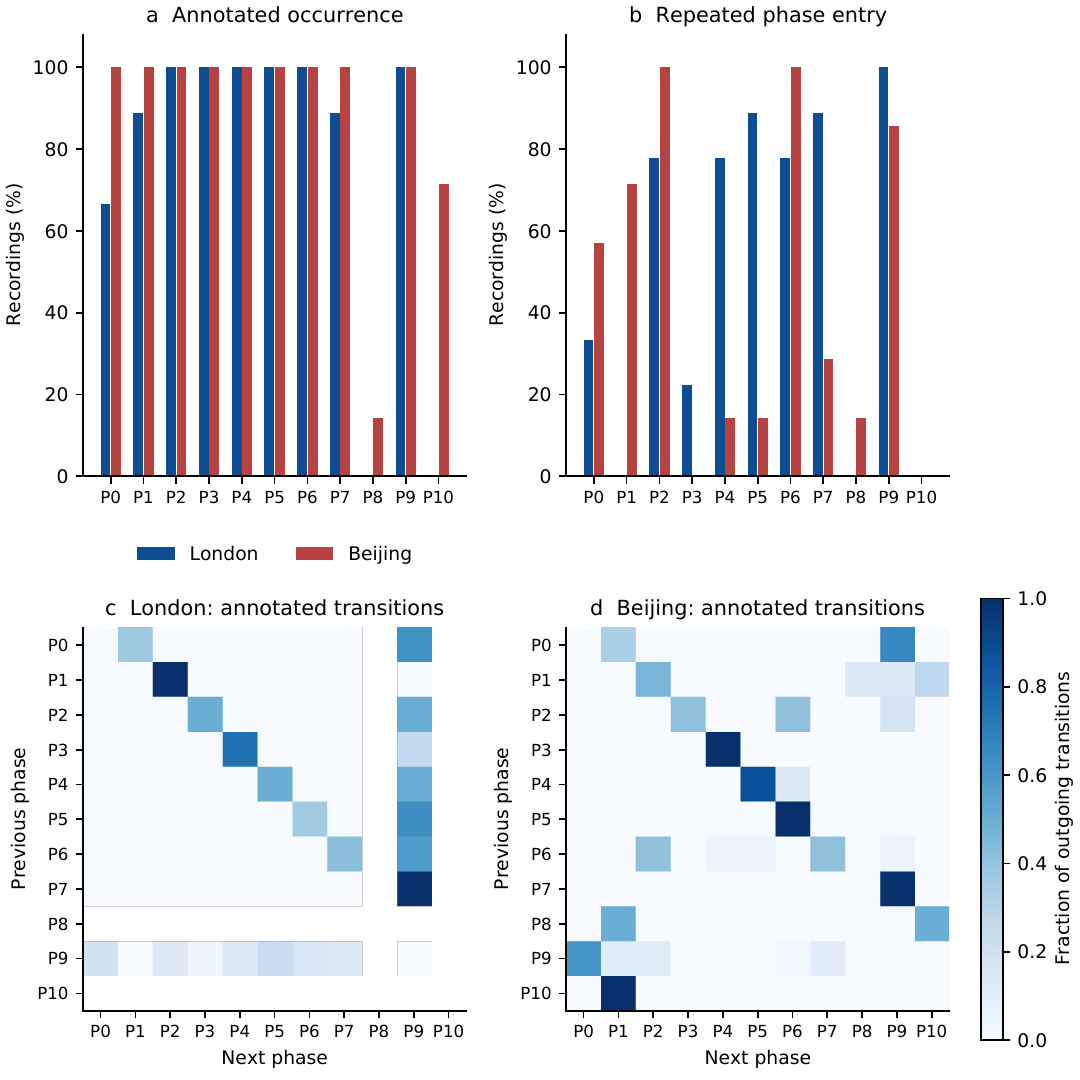}
\caption{Annotated workflows in London ($n=9$) and Beijing ($n=7$). Occurrence and re-entry are proportions of recordings. Transitions between adjacent runs are normalised by the outgoing counts of each phase; white rows have no transitions or lie outside the centre's schema. Phase definitions appear in Supplementary Table~\ref{stab:taxonomy}.}
\label{sfig:workflow}
\end{figure}

Supplementary Figs.~\ref{sfig:gt}--\ref{sfig:workflow} summarise the annotations of the evaluation recordings. Occurrence is the proportion of recordings that contain a phase, and re-entry is the proportion in which the phase occurs in more than one run of consecutive identical labels. Transition matrices count adjacent runs and are normalised by the outgoing counts of each phase; transitions across unlabelled intervals are not counted. A phase in the centre's schema that is absent from a recording contributes zero duration.

\clearpage
\clearpage
\clearpage
\subsection*{Phase-specific recognition and localisation}

\begin{table}[!htbp]\centering\small
\caption{Phase recall (\%) on the nine London recordings. Probability mean and confidence gate combine the Milan and Beijing experts.}
\label{stab:phase}
\begin{tabular}{lrrrrr}
\toprule
Phase & London reference & Milan & Beijing & Probability mean & Confidence gate\\\midrule
P0 & 92.07 & 95.56 & 70.73 & 96.22 & 96.13\\
P1 & 82.79 & 61.10 & 29.63 & 56.42 & 54.38\\
P2 & 95.82 & 54.98 & 97.39 & 93.89 & 93.73\\
P3 & 83.99 & 36.76 & 23.68 & 31.22 & 31.16\\
P4 & 91.84 & 49.44 & 13.25 & 26.36 & 26.10\\
P5 & 91.51 & 74.20 & 52.83 & 77.37 & 76.35\\
P6 & 79.46 & 57.31 & 43.26 & 55.02 & 53.58\\
P7 & 99.07 & 96.19 & 82.04 & 97.15 & 96.88\\
P9 & 96.09 & 91.33 & 95.82 & 96.36 & 96.36\\
\bottomrule\end{tabular}

\end{table}
\begin{table}[!htbp]\centering\small
\caption{Precision, recall and F1 (\%) pooled over the nine London recordings, and mean absolute error (MAE) of phase duration in minutes.}
\label{stab:extended}
\begin{tabular}{llrrrr}
\toprule
Pipeline & Phase & Precision (\%) & Recall (\%) & F1 (\%) & MAE (min)\\\midrule
\multirow{9}{*}{Milan} & P0 & 46.66 & 95.56 & 62.70 & 2.06\\
 & P1 & 59.00 & 61.10 & 60.03 & 0.48\\
 & P2 & 92.56 & 54.98 & 68.99 & 6.80\\
 & P3 & 69.43 & 36.76 & 48.07 & 1.46\\
 & P4 & 91.05 & 49.44 & 64.08 & 7.28\\
 & P5 & 63.61 & 74.20 & 68.50 & 6.18\\
 & P6 & 68.60 & 57.31 & 62.45 & 4.19\\
 & P7 & 90.99 & 96.19 & 93.51 & 2.70\\
 & P9 & 29.46 & 91.33 & 44.56 & 10.05\\
\hline
\addlinespace[0.5em]
\multirow{9}{*}{Beijing} & P0 & 89.81 & 70.73 & 79.13 & 0.59\\
 & P1 & 28.09 & 29.63 & 28.84 & 2.36\\
 & P2 & 62.87 & 97.39 & 76.41 & 9.40\\
 & P3 & 43.91 & 23.68 & 30.76 & 2.51\\
 & P4 & 99.74 & 13.25 & 23.40 & 13.81\\
 & P5 & 48.79 & 52.83 & 50.73 & 9.92\\
 & P6 & 53.09 & 43.26 & 47.67 & 8.10\\
 & P7 & 72.02 & 82.04 & 76.70 & 12.96\\
 & P9 & 69.40 & 95.82 & 80.50 & 1.82\\
 \hline
\addlinespace[0.5em]
\multirow{9}{*}{Probability mean} & P0 & 71.81 & 96.22 & 82.24 & 0.67\\
 & P1 & 85.36 & 56.42 & 67.93 & 0.91\\
 & P2 & 78.56 & 93.89 & 85.54 & 3.73\\
 & P3 & 57.90 & 31.22 & 40.57 & 2.24\\
 & P4 & 96.80 & 26.36 & 41.43 & 11.59\\
 & P5 & 59.83 & 77.37 & 67.48 & 6.72\\
 & P6 & 77.38 & 55.02 & 64.31 & 5.68\\
 & P7 & 83.63 & 97.15 & 89.88 & 6.32\\
 & P9 & 66.84 & 96.36 & 78.93 & 2.11\\
\bottomrule\end{tabular}

\end{table}
\begin{figure}[!htbp]
\centering\includegraphics[width=\linewidth]{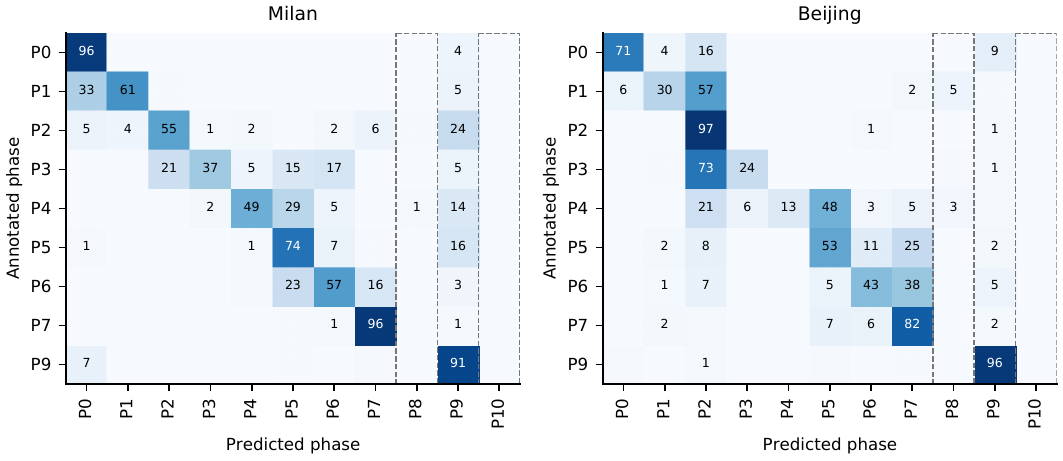}
\caption{Confusion matrices of the Milan and Beijing experts on the nine London recordings. Rows are GT phases and columns are predicted phases. Rows sum to 100\%, and values below 1\% are not printed. Dashed borders mark P8 and P10, which lie outside the London schema.}
\label{sfig:confusion}
\end{figure}
\begin{table}[!htbp]\centering\small
\caption{Segment metrics on the nine London recordings. Edit is averaged over recordings, and F1 pools segment matches at the stated intersection-over-union (IoU) thresholds.}
\label{stab:segment}
\begin{tabular}{lrrrr}
\toprule
Model / rule & Edit & F1@10 (\%) & F1@25 (\%) & F1@50 (\%)\\\midrule
Milan & 4.57 & 6.75 & 5.44 & 3.63\\
Beijing & 6.77 & 8.16 & 7.09 & 5.16\\
\addlinespace[0.5em]
Probability mean & 7.08 & 9.61 & 8.36 & 6.43\\
Confidence gate & 6.79 & 9.09 & 7.69 & 5.92\\
Entropy weighted & 7.08 & 9.61 & 8.35 & 6.42\\
\addlinespace[0.5em]
London reference & 29.85 & 37.32 & 37.13 & 33.11\\
\bottomrule\end{tabular}

\end{table}
\begin{table}[!htbp]\centering\small
\caption{Boundary localisation on the nine London recordings. Pred. and GT are the numbers of predicted and annotated boundaries. MAE and the percentage within 30~s are computed over matched boundaries.}
\label{stab:boundary}
\begin{tabular}{lrrrrr}
\toprule
Model / rule & Pred. & GT & Matched & MAE (s) & Within 30 s (\%)\\\midrule
Milan & 4,856 & 195 & 184 & 99.5 & 57.6\\
Beijing & 3,511 & 195 & 144 & 214.7 & 62.5\\
\addlinespace[0.5em]
Probability mean & 3,303 & 195 & 165 & 161.0 & 59.4\\
Confidence gate & 3,505 & 195 & 168 & 151.9 & 60.7\\
Entropy weighted & 3,306 & 195 & 168 & 156.3 & 60.1\\
\addlinespace[0.5em]
London reference & 832 & 195 & 182 & 61.4 & 83.0\\
\bottomrule\end{tabular}

\end{table}
\clearpage
\subsection*{Within-centre duration references}
Within each centre, the local reference estimated phase duration with a mean absolute error of 1.30 minutes in London, 1.85 in Milan and 1.14 in Beijing (main-text Fig.~\ref{fig:duration}). Across phases, the Spearman correlation between mean annotated duration and duration MAE was 0.60 in London, 0.32 in Milan and 0.56 in Beijing; the interval for Milan included zero.

\clearpage
\begin{figure}[htbp]
\centering\includegraphics[width=\linewidth]{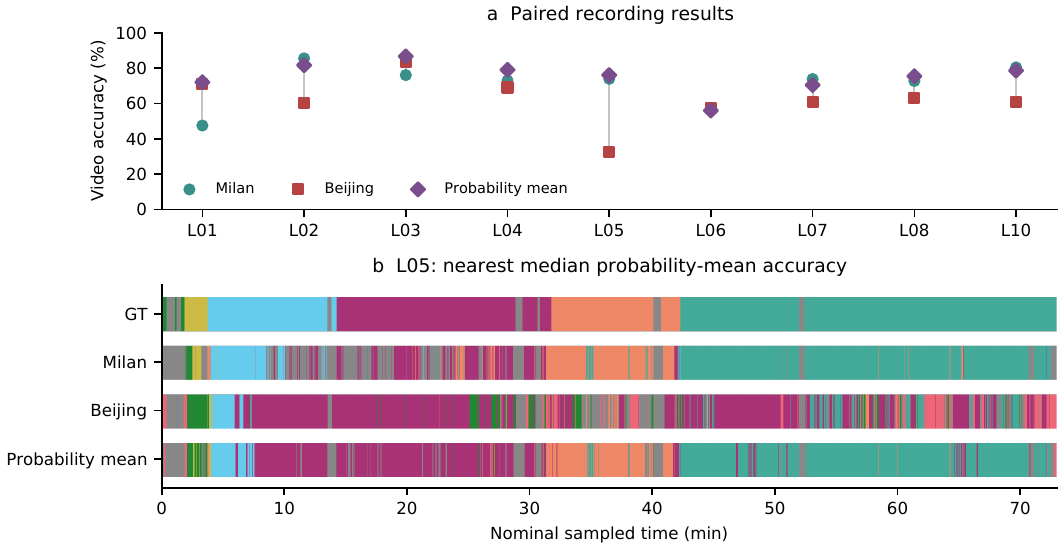}
\caption{Recognition in each London recording and an example timeline. (a) Accuracy in each of the nine London recordings; grey lines join the methods within a recording. (b) The recording nearest the median accuracy under probability averaging. Colours follow Supplementary Table~\ref{stab:taxonomy}.}
\label{fig:cases}
\end{figure}
\subsection*{Fixed fusion rules and subset comparisons}
For each expert $e\in\{\mathrm{Milan},\mathrm{Beijing}\}$, class probabilities were $\mathbf p_t^{(e)}=\operatorname{softmax}(\mathbf z_t^{(e)})$. Confidence gating took, at each sample, the prediction of the expert with the higher maximum probability,
\begin{equation}
e_t^*=\underset{e}{\arg\max}\,\max_c p_{t,c}^{(e)},\qquad
\hat y_t=\underset{c}{\arg\max}\,p_{t,c}^{(e_t^*)}.
\end{equation}
Ties went to the Milan expert. Probability averaging and entropy weighting instead combine the two probability vectors,
\begin{equation}
\mathbf p_t^{\mathrm{mean}}=\tfrac12\bigl(\mathbf p_t^{(M)}+\mathbf p_t^{(B)}\bigr),\qquad
\mathbf p_t^{\mathrm{ent}}=\sum_e w_t^{(e)}\,\mathbf p_t^{(e)},\quad
w_t^{(e)}\propto 1-\frac{H(\mathbf p_t^{(e)})}{\log 11},
\end{equation}
where $M$ and $B$ denote the Milan and Beijing experts and $H$ is the Shannon entropy. Weights were normalised to sum to one, and the fused prediction was the class with the highest combined probability.

\begin{table}[!htbp]\centering\small
\caption{Fusion rules on the nine London recordings. Frame, Phase and Video are percentages as defined in the main text, and duration MAE is in minutes. Intervals are 95\% bootstrap intervals over whole recordings.}
\label{stab:fusion}
\begin{tabular}{lrrrr}
\toprule
Model / rule & Frame (\%) [95\% CI] & Phase (\%) & Video (\%) & MAE (min)\\\midrule
Milan & 72.27 [64.25, 78.50] & 68.54 & 71.15 & 4.58\\
Beijing & 62.66 [55.63, 69.05] & 56.52 & 62.11 & 6.83\\
\addlinespace[0.5em]
Probability mean & 75.43 [69.93, 79.96] & 70.00 & 75.18 & 4.44\\
Confidence gate & 74.90 [69.39, 79.45] & 69.41 & 74.60 & 4.49\\
Entropy weighted & 75.26 [69.76, 79.80] & 69.80 & 75.01 & 4.46\\
\bottomrule\end{tabular}

\end{table}
\begin{table}[!htbp]\centering\small
\caption{Fusion rules within subsets of GT phases on the nine London recordings. Changes are relative to the Milan expert, in percentage points (pp) with paired 95\% bootstrap intervals over whole recordings. All includes P0, and P1--P7 does not.}
\label{stab:subsets}
\begin{tabular}{llrr}
\toprule
GT subset & Pipeline & Frame (\%) & Change vs Milan (pp) [95\% CI]\\\midrule
\multirow{5}{*}{All} & Milan & 72.27 & +0.00 [+0.00, +0.00]\\
 & Beijing & 62.66 & -9.61 [-19.49, +1.19]\\
 & Probability mean & 75.43 & +3.16 [-1.25, +9.15]\\
 & Confidence gate & 74.90 & +2.63 [-1.77, +8.60]\\
 & Entropy weighted & 75.26 & +2.99 [-1.43, +9.00]\\
\addlinespace[0.5em]
\multirow{5}{*}{P1--P7} & Milan & 70.95 & +0.00 [+0.00, +0.00]\\
 & Beijing & 60.97 & -9.97 [-20.44, +1.44]\\
 & Probability mean & 74.07 & +3.12 [-1.55, +9.50]\\
 & Confidence gate & 73.50 & +2.56 [-2.08, +8.91]\\
 & Entropy weighted & 73.88 & +2.94 [-1.74, +9.33]\\
\addlinespace[0.5em]
\multirow{5}{*}{P9} & Milan & 91.33 & +0.00 [+0.00, +0.00]\\
 & Beijing & 95.82 & +4.49 [-0.54, +8.26]\\
 & Probability mean & 96.36 & +5.03 [+0.32, +8.44]\\
 & Confidence gate & 96.36 & +5.03 [+0.32, +8.44]\\
 & Entropy weighted & 96.32 & +4.99 [+0.19, +8.41]\\
\bottomrule\end{tabular}

\end{table}
\clearpage
\section*{Supplementary Note 3: Inference latency, memory and numerical consistency}
\begin{figure}[!htbp]
\centering\includegraphics[width=\linewidth]{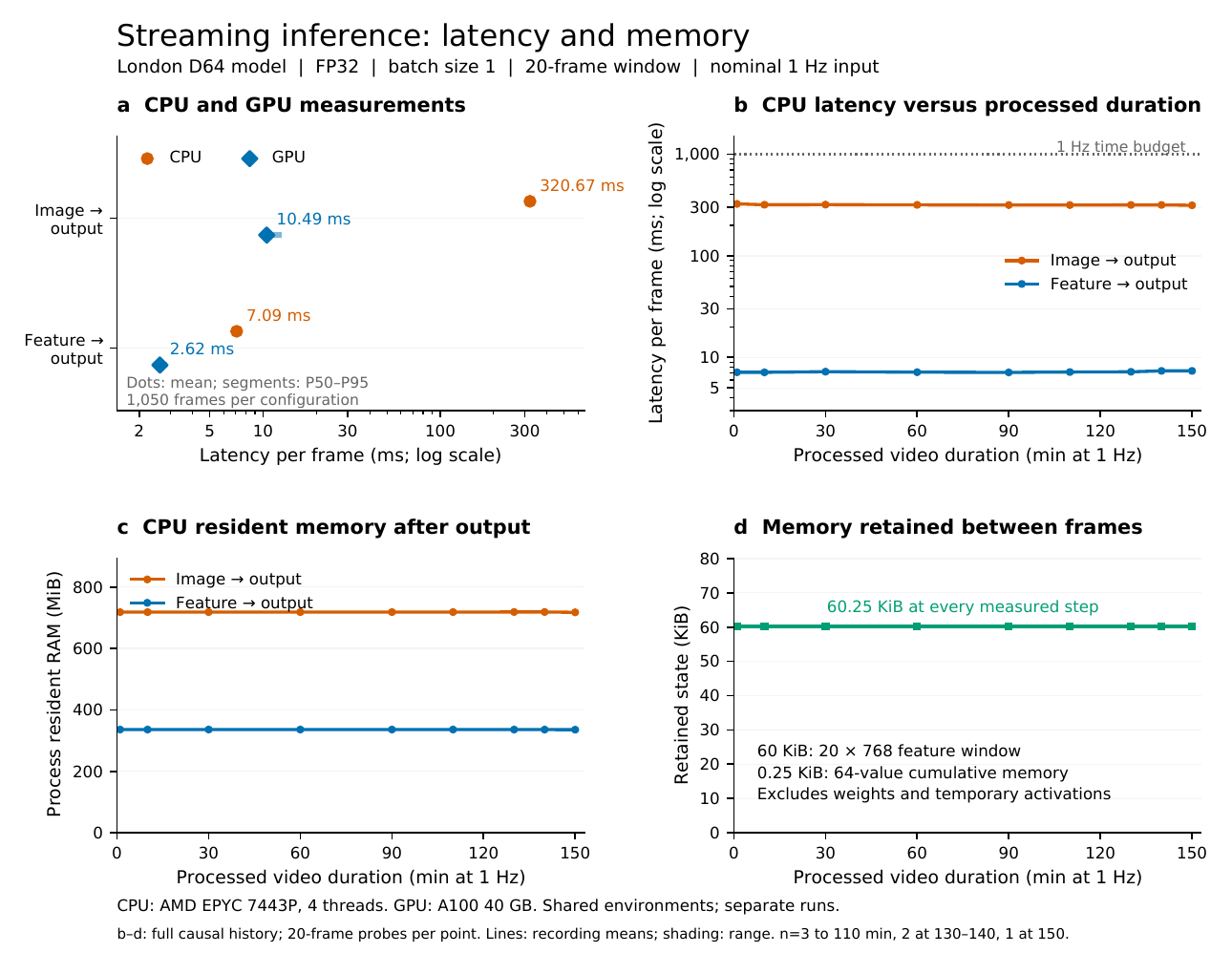}
\caption{Inference latency and memory as the processed duration increases. (a) Mean latency on CPU and GPU, with bars from the 50th to the 95th percentile (1,050 frames per configuration). (b) CPU latency at each probed duration; the dotted line marks the 1-s interval between frames. (c) Resident memory after each output. (d) State carried between frames, excluding weights and temporary activations. Lines are means over recordings and shading shows their range. Three recordings reach 110 minutes, two reach 130--140 minutes and one reaches 150 minutes.}
\label{fig:inference}
\end{figure}

\begin{figure}[!htbp]
\centering\includegraphics[width=\linewidth]{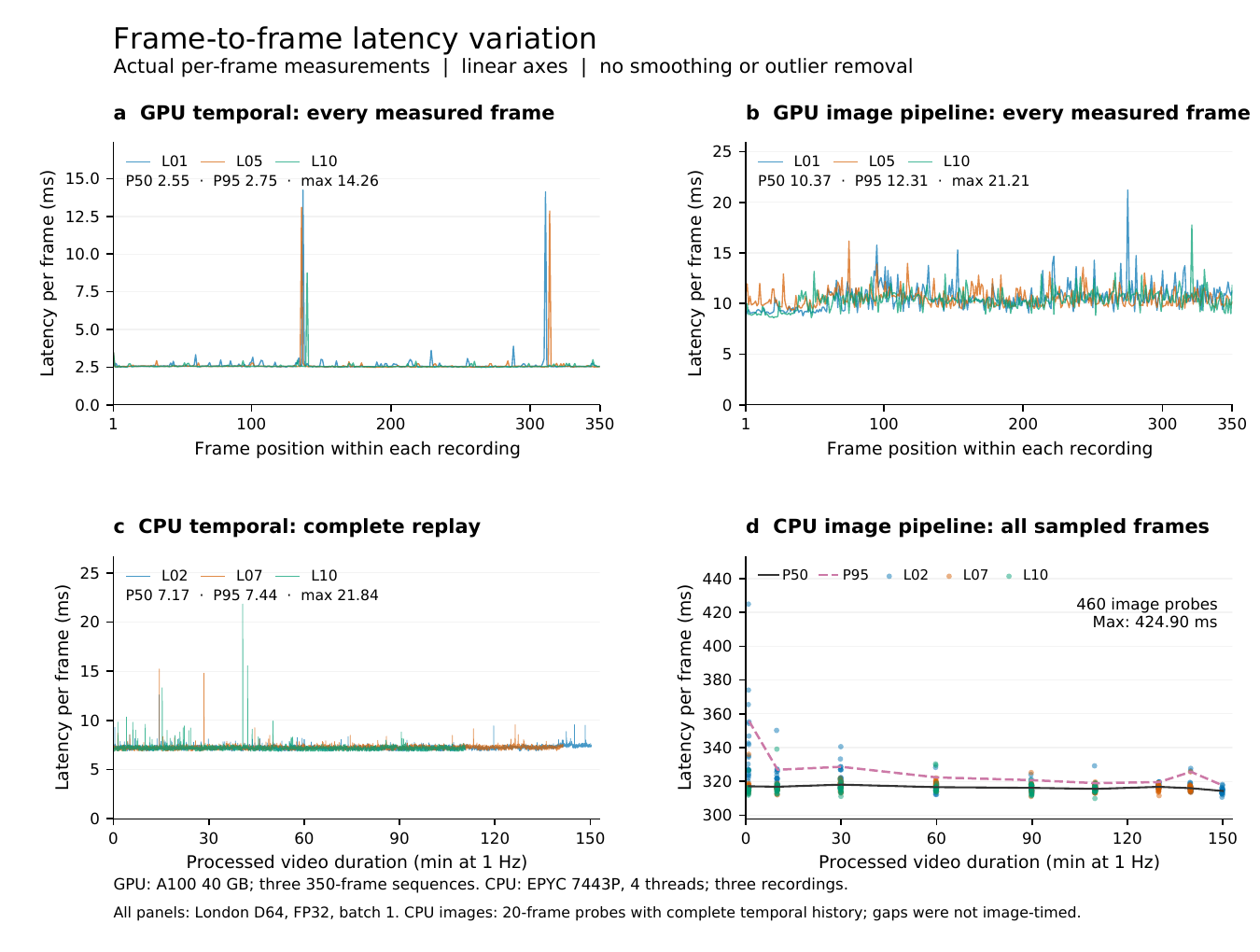}
\caption{Per-frame latency. (a) Temporal model and (b) full pipeline on the GPU for three 350-frame sequences. (c) Temporal model on the CPU over the complete recordings. (d) Full pipeline on the CPU at the probed frames, with lines joining the P50 and P95 values of each probe. Every measurement is shown.}
\label{sfig:inference-jitter}
\end{figure}

The London reference took a mean of 10.49~ms per frame on an NVIDIA A100 GPU, from image file to phase probabilities (95th percentile, 12.31~ms; maximum, 21.21~ms; 1,050 frames). On four CPU threads the mean was 320.67~ms. The temporal model alone took 2.62~ms on the GPU and 7.09~ms on the CPU (Supplementary Fig.~\ref{fig:inference}a and Supplementary Table~\ref{stab:inference-timing}). Single frames occasionally took longer (Supplementary Fig.~\ref{sfig:inference-jitter}).

Latency and memory did not grow with the length of video already processed. Between 1 and 150 minutes of video, CPU latency and resident memory showed no increase (Supplementary Fig.~\ref{fig:inference}b--d). The state carried between frames stayed at 60.25~KiB, of which the 20-sample feature window takes 60~KiB and the 64-value key memory 0.25~KiB. Frame-by-frame inference reproduced every predicted label of the three primary models on the nine London recordings.
\paragraph{Hardware and settings.}
CPU tests ran on an AMD EPYC 7443P with four threads (PyTorch 1.13.0), and GPU tests on an NVIDIA A100-SXM4-40GB (PyTorch 1.11.0, CUDA 11.3). Both used 32-bit floating point, a batch size of one and the 64-channel primary models, without inference optimisation or quantisation.

The latency benchmark used the first 350 frames of L01, L05 and L10 for each configuration. Latency from image file covers file reading, decoding, resizing, normalisation, visual encoding and temporal inference; it excludes camera acquisition, video decoding, transmission and display. The duration experiment replayed L02, L07 and L10 in full on the CPU. The full pipeline was timed on the last 20 frames before each of 1, 10, 30, 60, 90, 110, 130, 140 and 150 minutes.

\begin{table}[!htbp]\centering\small
\caption{Inference latency on recorded video. Panel (a) uses 350 consecutive frames from each of L01, L05 and L10. Panel (b) uses the complete recordings L02, L07 and L10 for the temporal model and the probed frames for the full pipeline. $n$ is the number of frames timed, and P50, P95 and P99 are latency percentiles. Rows with feature input time the temporal model alone; rows with image input time the full pipeline on London images.}
\label{stab:inference-timing}
\begin{tabular}{lllrrrrrr}\toprule
\multirow{2}{*}{Device} & \multirow{2}{*}{Input} & \multirow{2}{*}{Model} & \multirow{2}{*}{$n$} & \multicolumn{5}{c}{Latency (ms)}\\
\cmidrule(lr){5-9}
 & & & & Mean & P50 & P95 & P99 & Max.\\\midrule
\multicolumn{9}{l}{\textit{(a) Three 350-frame sequences per configuration}}\\[.35em]
\multirow{6}{*}{CPU} & \multirow{4}{*}{Features} & London & \multirow{6}{*}{1,050} & 7.09 & 7.08 & 7.38 & 7.57 & 8.14\\
 &  & Milan &  & 7.09 & 7.05 & 7.45 & 7.70 & 9.46\\
 &  & Beijing &  & 7.08 & 7.06 & 7.31 & 7.52 & 8.09\\
 &  & Probability mean &  & 14.73 & 14.72 & 15.15 & 15.44 & 15.80\\
\cmidrule(lr){2-3}\cmidrule(lr){5-9}
 & \multirow{2}{*}{Images} & London &  & 320.67 & 320.25 & 325.90 & 336.70 & 359.23\\
 &  & Milan &  & 332.13 & 329.30 & 364.62 & 396.95 & 435.51\\
\addlinespace[.45em]
\multirow{6}{*}{GPU} & \multirow{4}{*}{Features} & London & \multirow{6}{*}{1,050} & 2.62 & 2.55 & 2.75 & 3.24 & 14.26\\
 &  & Milan &  & 2.58 & 2.53 & 2.73 & 3.94 & 4.66\\
 &  & Beijing &  & 2.54 & 2.52 & 2.70 & 3.03 & 4.35\\
 &  & Probability mean &  & 4.97 & 4.93 & 5.18 & 5.81 & 8.95\\
\cmidrule(lr){2-3}\cmidrule(lr){5-9}
 & \multirow{2}{*}{Images} & London &  & 10.49 & 10.37 & 12.31 & 13.91 & 21.21\\
 &  & Milan &  & 8.95 & 8.81 & 9.69 & 10.89 & 16.62\\
\addlinespace[.45em]
\midrule\multicolumn{9}{l}{\textit{(b) CPU duration experiment: complete temporal replay and image probes}}\\[.35em]
\multirow{2}{*}{CPU} & Features & \multirow{2}{*}{London} & 24,142 & 7.18 & 7.17 & 7.44 & 7.66 & 21.84\\
 & Images &  & 460 & 318.13 & 316.51 & 327.35 & 351.88 & 424.90\\
\bottomrule
\end{tabular}

\end{table}

\clearpage

\clearpage

\section*{Supplementary Note 4: Key memory of the compact models}
Each of the eight channels of a compact model selects one frame per recording, which gives 72 selections in London, 80 in Milan and 56 in Beijing (208 in total). Milan images are the frames nearest to each selected sample in videos exported at about one frame every two seconds. The candidate scene descriptions (Supplementary Table~\ref{stab:hypotheses}) draw on published accounts of the technique\cite{huynh2018rarp} and were not annotated by clinicians.

\begin{figure}[!htbp]
\centering\includegraphics[width=\linewidth]{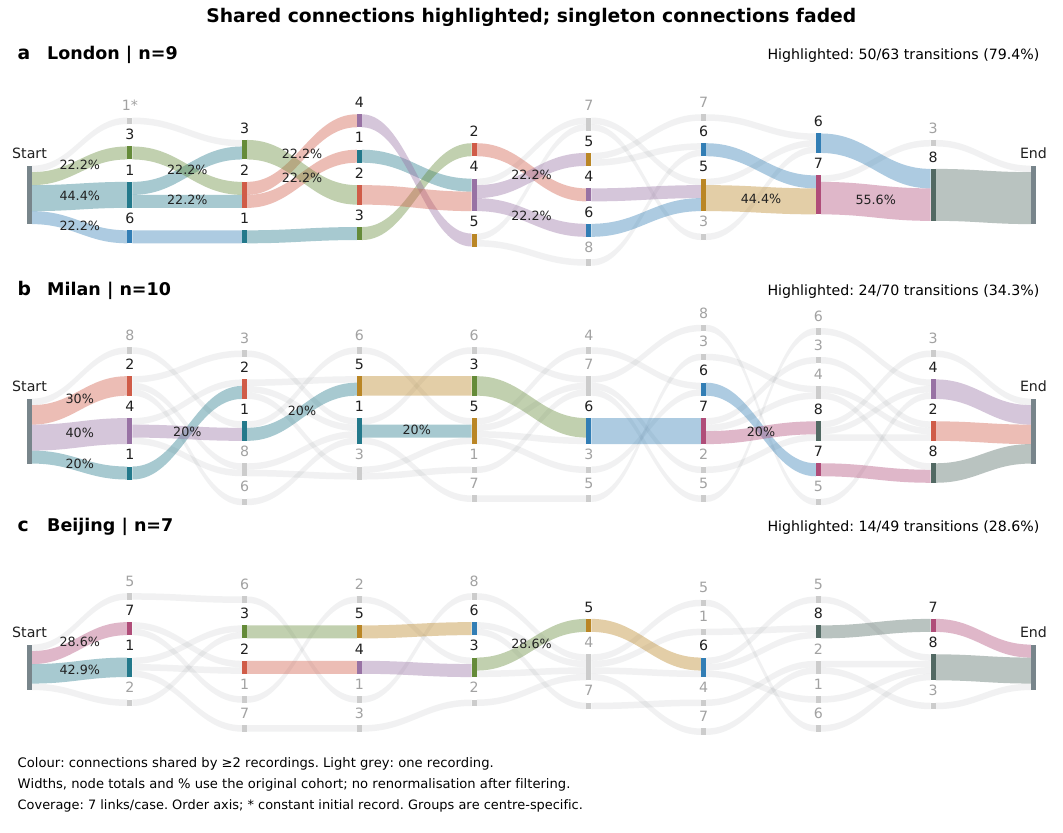}
\caption{Order in which the channels of the compact models reach their final maximum in (a) London ($n=9$ recordings), (b) Milan ($n=10$) and (c) Beijing ($n=7$). Columns give the order within a recording, and ribbon width is the number of recordings. Group numbers match main-text Fig.~\ref{fig:memory-scenes} within each centre and are not aligned between centres. Connections shared by at least two recordings are coloured, and those found in one recording are light grey. Percentages use all evaluation recordings of the centre as the denominator. The asterisk marks a London channel that kept its initial value throughout one recording.}
\label{sfig:channel-order}
\end{figure}

Agreement in channel order between recordings was measured with Kendall's $W$\cite{kendall1939} (Supplementary Table~\ref{stab:timing} and Supplementary Fig.~\ref{sfig:channel-order}). Connections between adjacent channels that were shared by at least two recordings made up 50 of 63 connections in London, 24 of 70 in Milan and 14 of 49 in Beijing.
\begin{table}[!htbp]\centering\small
\caption{Timing of the final maxima in the London and Beijing compact models. Kendall's $W$ measures agreement in channel order between recordings, and the leave-one-out minimum and maximum show its sensitivity to single recordings.}
\label{stab:timing}
\begin{tabular}{lrr}\toprule
Measure & London & Beijing\\\midrule
Recordings & 9 & 7\\
Selections & 72 & 56\\
Mean available span (min) & 109.0 & 101.5\\
Kendall's $W$ & 0.665 & 0.252\\
Leave-one-out minimum $W$ & 0.632 & 0.213\\
Leave-one-out maximum $W$ & 0.744 & 0.406\\
\bottomrule\end{tabular}

\end{table}
\begin{table}[!htbp]\centering\small
\caption{GT phases of the frames selected by the London and Beijing compact models. Groups follow the display order of each model and contain one selection per recording. U marks unlabelled samples, and Rep. GT is the phase of the displayed representative.}
\label{stab:groups}
\begin{tabularx}{\linewidth}{@{}lc>{\raggedright\arraybackslash}X>{\raggedright\arraybackslash}p{6.5cm}c@{}}\toprule
Centre & Group & Candidate scene & GT counts & Rep. GT\\\midrule
\multirow[t]{8}{*}{London} & 1 & Mixed views & P0: 3; P1: 2; P2: 3; P9: 1 & P2\\
 & 2 & Broad membrane plane & P2: 8; P3: 1 & P2\\
 & 3 & Tented thin tissue & P1: 7; P6: 2 & P1\\
 & 4 & Pale longitudinal surface & P3: 9 & P3\\
 & 5 & Opening / pale structures & P3: 2; P4: 4; P5: 3 & P4\\
 & 6 & Irregular exposed surface & P0: 2; P4: 4; P6: 3 & P6\\
 & 7 & Narrow tissue cleft & P5: 8; P6: 1 & P5\\
 & 8 & Deep cavity / fine strands & P4: 1; P7: 8 & P7\\
\addlinespace[0.5em]
\multirow[t]{8}{*}{Beijing} & 1 & Mixed / degraded views & P5: 2; P6: 1; P9: 4 & P5\\
 & 2 & Yellow tissue window & U: 1; P5: 1; P10: 5 & P10\\
 & 3 & Fat and membrane plane & U: 1; P1: 1; P2: 4; P8: 1 & P2\\
 & 4 & Central tissue recess & U: 1; P4: 1; P5: 1; P6: 1; P10: 3 & P10\\
 & 5 & Pale oval surface & P3: 6; P4: 1 & P3\\
 & 6 & Pale tube in an opening & P3: 1; P4: 6 & P4\\
 & 7 & Smooth tissue / occlusion & U: 1; P0: 3; P4: 1; P7: 2 & P0\\
 & 8 & Deep cavity / fine strands & P7: 7 & P7\\
\bottomrule\end{tabularx}

\end{table}
\begin{table}[!htbp]\centering\small
\caption{Candidate interpretations of the selected scenes and the checks they require. The interpretations are hypotheses that need clinical review of the surrounding video.}
\label{stab:hypotheses}
\begin{tabularx}{\linewidth}{>{\raggedright\arraybackslash}p{3.0cm}>{\raggedright\arraybackslash}X>{\raggedright\arraybackslash}X}\toprule
Visible context & Possible relevance to historical context & Specific check before clinical interpretation\\\midrule
Broad exposed surfaces or a tissue plane held under traction & Earlier field geometry could help distinguish similar later close-up views. Current traction alone may need only a short window. & Confirm the anatomical compartment in a clip; compare with nearby, non-selected frames. Appearance alone does not establish a named fascial plane.\\[.6em]
An opening with a pale or tubular structure & A confirmed landmark could contextualise an earlier opening or exposure. & Verify whether the structure is anatomy, catheter or another object. Do not distinguish bladder neck from apex from an isolated still.\\[.6em]
Deep operative cavity with visible suture-like material & Could identify a reconstructive setting, while current needle handling may be locally sufficient. & Confirm visible material and action in a clip. P7 also includes posterior reconstruction and haemostasis sutures, so suturing alone does not identify anastomosis.\\[.6em]
Mixed, obscured or extracorporeal views & No specific operative-context role assigned; these may be nuisance responses or recording-state cues. & Check consistency across cases and test whether obscuration, colour or camera changes explain the selection.\\\bottomrule
\end{tabularx}
\end{table}
London groups 2 and 3 selected P2 and P1 in 88.9\% and 77.8\% of recordings, although these phases make up 14.7\% and 1.4\% of samples. Beijing groups 2, 5 and 6 selected P10, P3 and P4 in 71.4\%, 85.7\% and 85.7\% of recordings, against 5.3\%, 2.4\% and 9.1\% of samples (Supplementary Table~\ref{stab:groups}).

\end{document}